\documentclass[letterpaper, 10 pt, conference]{ieeeconf}  % Comment this line out if you need a4paper

\IEEEoverridecommandlockouts                              % This command is only needed if 
\usepackage{graphics} % for pdf, bitmapped graphics files
\usepackage{epsfig} % for postscript graphics files
\usepackage{mathptmx} % assumes new font selection scheme installed
\usepackage{times} % assumes new font selection scheme installed
\usepackage{microtype}
\usepackage{amsmath,amssymb,amsfonts}
\usepackage{algorithmic}
\usepackage{textcomp}
\usepackage{xcolor}
\usepackage{gensymb}
\usepackage{caption}
\usepackage{hyperref}
\usepackage{mwe}
\usepackage{verbatim}
\usepackage{amsbsy}
\usepackage{siunitx}
\usepackage{tabularx, booktabs}
\usepackage{dblfloatfix}
\usepackage{cite}
\usepackage{bm}
\usepackage{float}
\usepackage{subcaption}
\usepackage{algorithm, algorithmic}

\title{\LARGE \bf
Depth-CUPRL: Depth-Imaged Contrastive Unsupervised Prioritized Representations in Reinforcement Learning for Mapless Navigation of Unmanned Aerial Vehicles
}

\author{Junior C. de Jesus$^{1}$, Victor A. Kich$^{2}$, Alisson H. Kolling$^{2}$, \\Ricardo B. Grando$^{3}$,  Rodrigo S. Guerra$^{2}$, Paulo L. J. Drews-Jr$^{1}$% <-this % stops a space
\thanks{$^{1}$Junior C. de Jesus  and P. L. J. Drews-Jr are with the NAUTEC, Centro de Ciencias Computacionais, Universidade Federal do Rio Grande - FURG, RS, Brazil.
        {\tt\small dranaju@gmail.com}}
\thanks{$^{3}$Victor A. Kich, Alisson H. Kolling and Rodrigo S. Guerra are with the Universidade Federal de Santa Maria - UFSM, RS, Brazil.
        {\tt\small rodrigo.guerra@ufsm.br}}%
\thanks{$^{3}$Ricardo B. Grando is with Technological University of Uruguay, Rivera, Uruguay
        {\tt\small bedingrando@gmail.com}}%
}

\begin{document}

\maketitle
\thispagestyle{empty}
\pagestyle{empty}
\begin{abstract}

Reinforcement Learning (RL) has presented an impressive performance in video games through raw pixel imaging and continuous control tasks. However, RL performs poorly with high-dimensional observations such as raw pixel images. It is generally accepted that physical state-based RL policies such as laser sensor measurements give a more sample-efficient result than learning by pixels. This work presents a new approach that extracts information from a depth map estimation to teach an RL agent to perform the mapless navigation of Unmanned Aerial Vehicle (UAV). We propose the Depth-Imaged Contrastive Unsupervised Prioritized Representations in Reinforcement Learning(Depth-CUPRL) that estimates the depth of images with a prioritized replay memory. We used a combination of RL and Contrastive Learning to lead with the problem of RL based on images. From the analysis of the results with Unmanned Aerial Vehicles (UAVs), it is possible to conclude that our Depth-CUPRL approach is effective for the decision-making and outperforms state-of-the-art pixel-based approaches in the mapless navigation capability.

\end{abstract}

\section*{SUPPLEMENTARY MATERIAL}

Video of the experiments is available at: \url{https://youtu.be/-iNnP3HLXmI}. Released code at: \url{https://github.com/dranaju/curl_navigation}.
\section{INTRODUCTION}\label{introduction}

%%%%%%%%%%%%%%%%%%%%%%%%%%%%%%%5555

% Obstacle avoidance is an essential feature for an UAV to navigate autonomously in various types of complex environments \cite{radmanesh2018overview}. 
The problem of autonomous robotics navigation can be defined in the idea of perceiving the environment and planning a path that leads a robot to avoid places that may represent a specific danger\cite{alves2011conceptual}. This first principle can be solved by using senses to tell the robot's environment; examples are laser sensors, cameras, or depth maps. In the second principle for an autonomous vehicle, it is necessary to use systems to plan the robot's path to avoid dangerous situations, such as collisions and unsafe conditions. Linking these principles of navigation problems to UAVs, it is possible to see the great difficulty found in contrasting means of performance and perception \cite{grando2020visual}. %To deal with this problem \cite{ricardo2021aprend} proposed an intelligent reinforcement planning system using laser perceptions.

% Deep Reinforcement Learning (RL) is a promising approach to tackle the problem of autonomous robotics navigation. In this so called mapless navigation \cite{tai2016towards}, and RL agent learns good behavior \cite{franccois2018introduction} by modifying or acquiring new behaviors and skills incrementally. Another critical aspect of RL is that it uses experience from trial and error. Furthermore, the RL agent does not need complete knowledge or control of the environment, and it just needs to interact with the environment and collect information. An agent performs specific actions in an environment, and an interpreter gives a reward for the new state of the agent, thus having a cycle between the actions and interpretations made in the environment. 
Deep Reinforcement Learning (Deep-RL) is a promising approach to tackle the problem of autonomous robotics navigation. In this so-called mapless navigation \cite{tai2016towards}, the methods are capable of providing an action response for continuous systems \cite{lillicrap2015continuous,haarnoja2018applied}, as well as the Deep-RL control of UAVs in simulated environments \cite{grando2020deep}. It is possible to see the application of these techniques for the mapless navigation of UAVs in models of unknown dynamics in many environments\cite{rodriguez2018deep, sampedro2019fully, he2020integrated, grando2022double, grando2020deep}. However, one of the significant challenges of Deep-RL techniques is the limitation given by the immense need for labeled data which is difficult to collect in the real world \cite{bonatti2019learning}. It has been empirically observed that high-dimensional Deep-RL observations with images of raw pixels are inefficient in terms of sample \cite{lake2017building}. Knowing this inefficiency case in RL, better techniques to extract information from images are required. 

\begin{figure}[t]
\includegraphics[width=\linewidth]{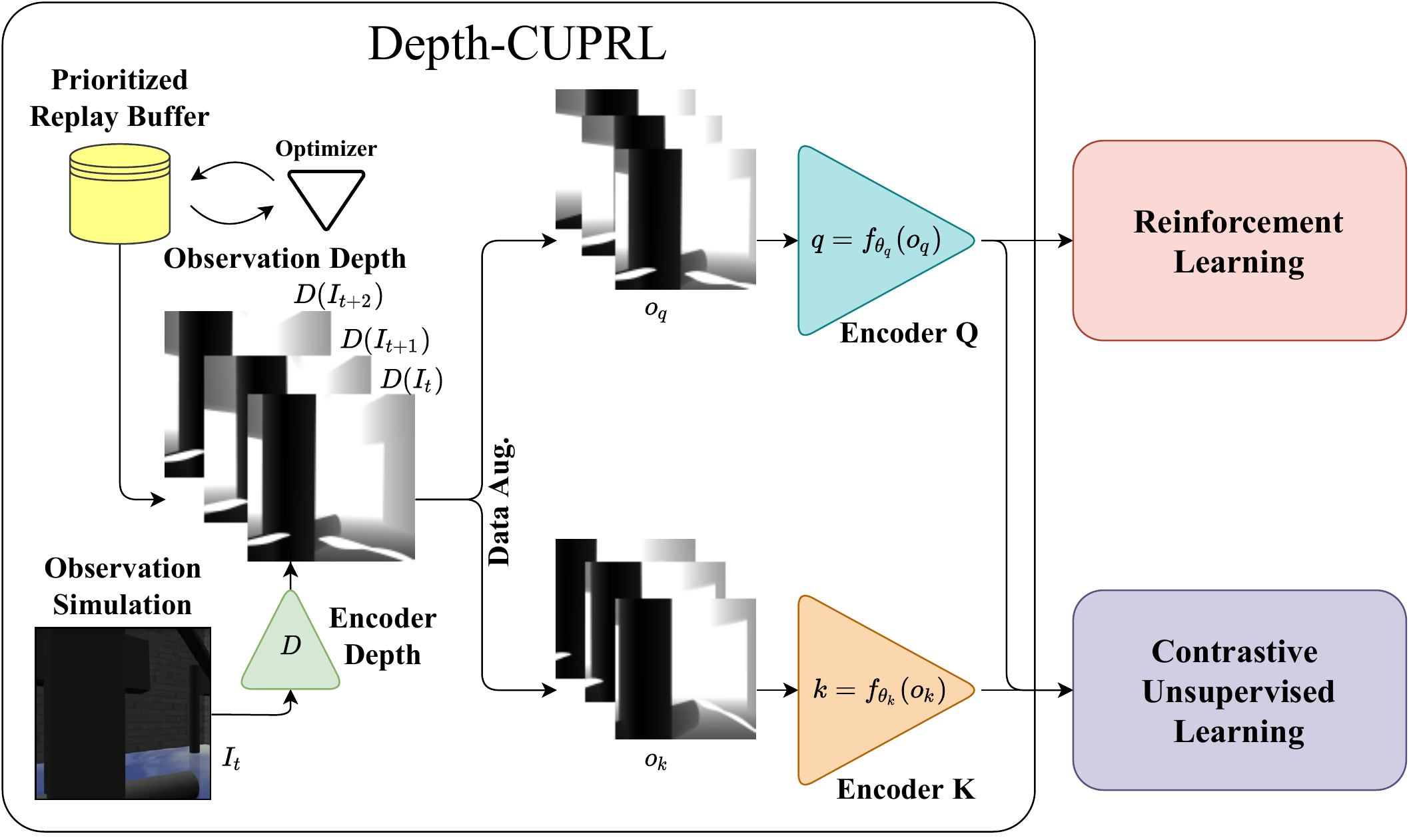}
\caption{System of Depth-CUPRL: The observations are passed through a network that generates the depth images. These depth images are stored in a replay memory, and it is sampled memories in the buffer to make the Contrastive and Reinforcement Learning.}
\label{fig:state_action}
\vspace{-6mm}
\end{figure}

Since depth maps are a well-known source of perception data for navigation, we propose a Deep-RL system to perform the mapless navigation of an UAV through features extracted from a depth map image. We propose the Depth-Imaged Contrastive Unsupervised Prioritized Representations in Reinforcement Learning (Depth-CUPRL) that considers information from the depth of images with a prioritized replay memory to perform the navigation of an UAV. We show that a combination of Deep-RL and Contrastive Learning leads to better performance when compared with other approaches of Deep-RL based on images. Fig. \ref{fig:state_action} shows the system of states $D(I_t,I_{t-1},I_{t-2})$ that will be used in networks for training and its architecture.

The main contributions of this work are listed as follows:

\begin{itemize}

\item We present a novel methodology based on depth images and contrastive learning that can successfully perform mapless navigation and obstacle avoidance for an UAV that manages to outperform image-based Deep-RL algorithms;

% \item We show that our agent is capable of generalizing by acting in an third unknown environment.
\item We also show that taking into account information from \textbf{depth images we are able to increase the performance by 43.9\% in success rate in the best case (ours)} and by 26.9\% in the worst depth-based method used by comparison. 

\item We also show that \textbf{prioritized experience replay} memory further enhances contrastive learning and generalization ability in deep-RL networks,\textbf{ increasing the performance by 5.6\% in success rate}. 

\end{itemize}

%%%%%%%%%%%%%%%%%%%%%% escrever por ultimo essa parte

The paper is divided as follows: Section \ref{related_works} presents a review of the relevant works in the field. After that, Sections \ref{theoretical} and \ref{methodology} present details related to the proposed methodology as well as a discussion of the developed tools. Section \ref{results}, the description of the vehicle and the environment are made and the results obtained are presented. Finally, our contribution and future works are summarized in Section \ref{conclusions}.
\section{RELATED WORKS}\label{related_works}

The study around mapless navigation is extensively explored using terrestrial mobile robots \cite{tai2016towards}. Autonomous navigation of aerial mobile robots using Deep-RL approaches is less frequent and mainly focused on approaches that avoid the use of visual information \cite{grando2022double, grando2020deep} or using simplified information, without contrastive learning \cite{rodriguez2018deep, sampedro2019fully, he2020integrated, li2020uav}.
% DREWS Some studies have already investigated the use of reinforcement learning for navigation in Autonomous Underwater Vehicles (AUV), such as Carreras \emph{et al.} \cite{carreras2001hybrid}, who proposed a hybrid coordination method for the control of behavior-based architectures. They show the feasibility of applying this hybrid method with 3D navigation in an AUV, where behaviors are learned online through reinforcement learning. In realistic simulations, a continuous action Q-learning algorithm is implemented with a feed-forward neural network. The obtained results show the excellent performance of the hybrid method in behavioral coordination and the convergence of behaviors.

Rodriguez \emph{et al.} \cite{rodriguez2018deep} implemented a Deep-RL approach to solve the problem of landing a UAV on a base that could be in motion or static. Sampedro \emph{et al.} \cite{sampedro2019fully} suggested that an approach based on the Deep-RL technique could perform a task of Search and Rescue in closed scenarios. They are modified through a Deep Deterministic Policy Gradient (DDPG) \cite{lillicrap2015continuous}. In the work of \cite{de2021soft}, it was shown that approaches based on the Soft Actor-Critic (SAC) algorithm \cite{haarnoja2018applied} is more effective for robot navigation than approaches based on the DDPG algorithm.

%DREWS Humans are remarkably capable of inferring depths and the 3D structure of a scene, even on short time scales. Taking this as a hypothesis, Zhou \emph{et al.} \cite{zhou2017unsupervised} presented an unsupervised learning framework for the task of estimating depth maps and camera movements of unstructured videos in sequence using only a monocular camera. Through depth maps, it is possible to navigate an underwater robot, as explored in the work of Drews-Jr \emph{et al.} \cite{drews2016real}, who estimates depth maps using the UDCP technique \cite{drews2013transmission} and controls an AUV to avoid obstacles by identifying regions of interest that allow you to calculate the escape direction and vehicle control.

Knowing that Deep-RL from high-dimensional observations (state) with raw pixel images is inefficient in terms of sample \cite{kaiser2019model}, Laskin \emph{et al.} \cite{laskin2020curl} presented a technique capable of extracting high-level characteristics from images with raw pixels for the control of Deep-RL networks through these extracted features. In addition, the CURL method showed a better performance than the state-of-the-art algorithms such as SAC when dealing with images. However, CURL is a method that was only applied in game-like environments with a third-person view in the environment.

For navigation for mobile aerial vehicles, it is possible to find works like Thomas \emph{et al.} \cite{thomas2021interpretable} which presents an algorithm for an autonomous UAV using RL with self-attentive models. It showed that it can effectively complete the vehicle navigation even when subjected to varying inputs in the algorithm. This work highlights how the navigation works with the input state data with noise or modifications.
He \emph{et al.} \cite{he2020integrated} used a Lobula Giant Moment Detector (LGMD) to simplify vision information for Deep-RL in navigation and obstacle avoidance of UAV. It performed missions in a complex environment with $80\%$ of success rate.

% The proposed work focuses on learning the depth map of aerial environments by unsupervised techniques through sequences of image frames captured by a monocular camera for the control of UAVs through algorithms Deep-RL that performs the actions. For this, it is necessary first that this technique be efficient in simulation environments, thus navigating safely in aerial environments and identifying everything that could be dangerous for the agent.

The present work differs from the related works by using information from a depth image in a contrastive learning-based approach. Our distance sensing information is yielded from a neural network capable of depth estimation in monocular images instead of a distance sensor such as a Lidar. We also differ by avoiding the high-dimensional observations problem by presenting a contrastive learning structure. Also, we outperform the state-of-the-art by proposing a prioritized experience replay memory system that further enhances our approach. To the best of our knowledge, we are the first to propose such a contrastive learning system with prioritized experience replay.

% follows the state-of-the-art work in mapless navigation for an UAV proposed by Grando \emph{et al.} \cite{bedin2021deep, grando2022double}. They explore two state-of-the-art Deep-RL techniques for navigation of a simulated UAV, taking into account only information from range-based sensors for decision making. Unlike the work of Grando \emph{et al.} \cite{ricardo2021aprend}, the purpose of this study is to explore techniques that can extract information through images since Deep-RL presents a poor learning process when dealing with this type of high-dimensional observation, which are raw pixel images. This work aims to improve the capability to extract information from RGB images, since most of the UAV available in the market presents a monocular camera, but it is very uncommon to have LIDAR or other range-based sensor.
\section{THEORETICAL BACKGROUND}\label{theoretical}

This section details the information of tools and techniques used to develop our work. 

\subsection{SAC-Based Approach}

% \cbox{ta muito mimimi, nao explica como foi feita a proposta, foca no que tu faz de diferente e porque tu "escolheu" alguma abordagem}

The main idea of the SAC technique is to use approximation functions that can learn policies of continuous action space, characterizing this method as a stochastic actor-critic \cite{haarnoja2018applied}.
% Therefore, a large continuous domain requires that we derive a practical approximation for a soft policy iteration.
% Soft policy iteration is an algorithm that learns optimal policies of maximum entropy and switches between policy evaluation and improvement.
% The policy evaluation step tries to find the proper $V$ value function for our current policy according to the maximum entropy.
% The policy improvement step updates the policy distribution towards the exponential distribution for the current $Q$ function.
The SAC algorithm uses a neural network to approximate a policy to the actor-network. The state value is used to estimate a value network $V$ and a value $Q$ with a critic network.
These three networks calculate the action prediction for the current state and generate a time difference error signal for each time interval step. In addition, intending to seek the maximization of the system's rewards, the SAC also seeks to maximize the entropy of the policy. Entropy refers to how unpredictable a variable can be. If a random variable still assumes a single value, then the strategy has zero entropy, encouraging exploration by the agent.

In general, 
% to train and evaluate the policy function given by the actor network, the value function, and the $Q$ function given by the critic network, 
it is recommended to make use of a replay memory that can save the experiences of the agent during the \cite{schaul2015prioritized} training session.
% It is necessary to save experiments because there are thousands of simulated temporally correlated trajectories that can accumulate a large number of variances when approaching the most qualified $Q$ function of the problem.
The states, actions, rewards, and new states that an agent has experienced in previous episodes are saved in a memory \textit{buffer}, being marked as an experience.
% When a minimum number of experiences are stored in memory, those experiences in previous episodes are randomly sampled for the learning algorithm.
This replay memory technique can disrupt temporal correlations between different episodes in the training phase.
A significant performance increase in off-policy Deep-RL algorithms is noticeable even with short memories.
Regardless of the increase in application capabilities with Deep-RL, replay memories can affect an agent's learning time.

Despite the use of replay memories, 
% \cbox{???Q-learning algorithms make use of spinning up \cite{spinningup2018} target network}.
% The expression uses the term target because minimizing the loss makes the $Q$ function closer to its target. However, the target depends on precisely the same parameters as the $\theta$ training. $\theta$ represents the weights of the main network. This case makes the minimization loss unstable.
% So the solution was to use 
another network for the target is used with a time delay, called the target network.
The update of target network weights can be defined as:
\begin{equation}
     \theta' \leftarrow \tau \theta + (1-\tau)\theta'
\end{equation}
with $\tau \ll 1 $.
This equation tells us that the target network is just a copy of the main network using a Poliak-averaging \cite{polyak1992acceleration} to update its parameters that affect the learning stability of Deep-RL algorithms.

The SAC technique was used in the CURL as the learning method, and we used it in our Depth-CUPRL approach as well. We also used this technique in a CNN-based approach with the SAC algorithm (SAC CNN Prio.) to further enhance the comparison with our proposed system.

\subsection{Contrastive Unsupervised Representations for Reinforcement Learning}

Contrastive Unsupervised Representations for Reinforcement Learning (CURL) is a general framework for combining contrastive learning with Deep-RL \cite{laskin2020curl}. In principle, any RL algorithm in the CURL pipeline can be used, either on-policy or off-policy. The SAC technique is primarily used in this case \cite{laskin2020curl}. CURL has been shown to significantly improve sample efficiency over previous pixel-based methods by performing contrastive learning simultaneously with an off-policy RL algorithm.

\begin{figure}[t]
\centering
\includegraphics[width=0.72\linewidth]{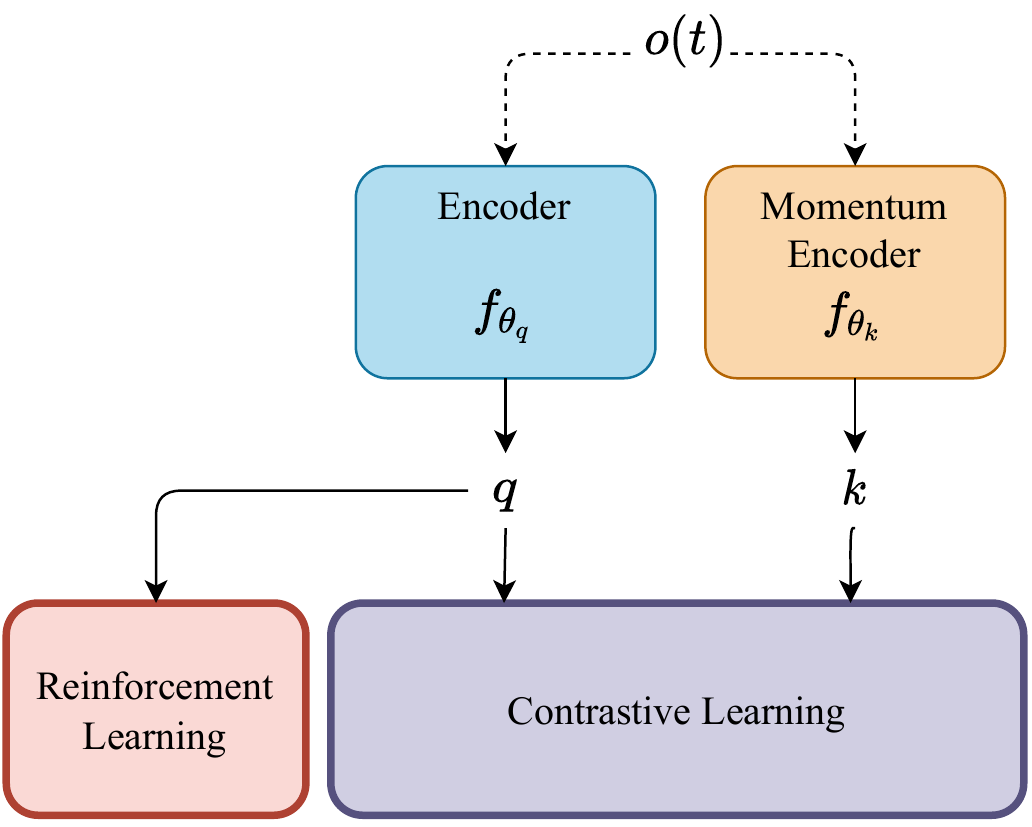}
\caption{CURL structure combines instance Contrastive Learning and Reinforcement Learning.}
\vspace{-5mm}
\label{fig:curl_struct}
\end{figure}

A key component of CURL is the ability to learn rich representations of dimensional data using contrastive unsupervised learning. Given a \textit{query} $q$ and \textit{key} $k$, the objective of contrastive learning is to secure that $q$ matches with $k$.
A \textit{query} and \textit{key} are positive pairs if they are modified data from the same observation $o(t)$, like an image observation. Fig. \ref{fig:curl_struct} shows that CURL trains a visual representation $encoder$, ensuring that embeddings of augmented versions of data $o_q$ and $o_k$ from observation $o(t)$ match using a contrastive loss of the work from Van der Oord \emph{et al.} \cite{oord2018representation}.

The observations from the \textit{query} $o_q$ are handled by \textit{encoder} $f_{\theta_q}$, returning $q=f_{\theta_q}(o_q)$. All observations are built from the minibatch ($i$) sampled from replay memory for the Deep-RL update. The \textit{key} is encoded and return $k=f_{\theta_k}(o_k)$, where $\theta_k$ is a Poliak-averaging version of the \textit{encoder query} or $\theta_k \leftarrow \tau \theta_k + (1-\tau)\theta_q$. The Deep-RL policy and value function are built on top of the \textit{query} encoder that is trained together with the objectives of reinforcement learning and contrastive learning:

\begin{equation}
    \mathcal{L}oss_{\theta_q} = \log \frac{\exp(q^Tk_0)}{ \sum\nolimits_{0}^{i}\exp(q^Tk_i) } 
\end{equation}

We further enhanced the CURL approach and added a prioritized replay memory system to create ours. We implemented a version of the CURL classic and a version with Depth Maps for a more extensively comparison. Both prioritized replay memory system and unsupervised depth estimation techniques are described below.

\subsection{Prioritized Replay Memory}

In the replay of experience, the agent can use a replay memory that allows data efficiency by storing past experiences of the agent to have the opportunity to reprocess the data later. The data saved in memory refers to the current state $s_t$, action $a_t$, reward $r_t$ and next state $s_{t+1}$ of the agent.
Furthermore, replay memories also ensure that updates are made from reasonably stable data stored in memory which helps with network convergence.

\begin{figure}[t]
    \includegraphics[width=\linewidth]{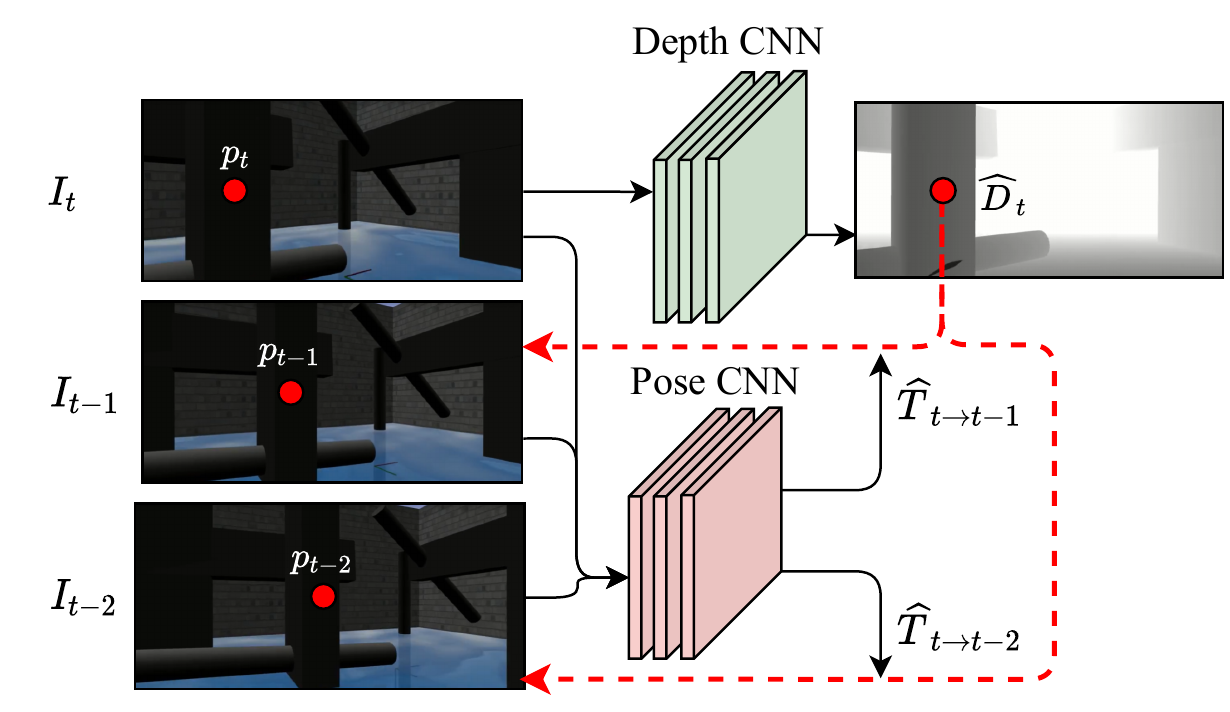}
    \caption{Unsupervised Depth estimation architecture. The depth network takes the current as input and outputs a depth map $\widehat{D_t}$. The pose network assists in the loss for the depth map estimation.}
    \label{fig:depth_and_pose_cnn_proposed_arch}
    \vspace{-5mm}
\end{figure}

While replay memory allows processing data transitions in different orders than if they were experienced, there is also the possibility to use \cite{schaul2015prioritized} prioritized memories. These prioritized experiences allow data transitions with frequencies different from those experimentally tested depending on their importance.
That is to say, which experience is stored and which one is repeated.
% The disadvantage of this prioritized repetition method is the introduction of trends.

We added a prioritized replay memory system to our approach and to some of the approaches that we used compared to. The idea is that such a prioritized system increases the ability of the Deep-RL algorithm to learn the depth representation.

\subsection{Unsupervised Depth Estimation}

One hypothesis is that the human being developed a rich, structural understanding of the world through our previous visual experience, which consisted mainly of moving around and observing large numbers of scenes and developing consistent models of our observations. From millions of such observations, we learn about the regularities of the world - roads are flat, buildings are straight, roads support cars, etc. We can apply this knowledge to perceiving a new scene, even from a single monocular image \cite{yin2018geonet}.

Given this, works such as Bian \emph{et al.} \cite{bian2021unsupervised} explore learning to estimate depth maps through image sequences. His approach is based on the insight that a geometric view synthesis system performs consistently well when its intermediate predictions of scene geometry and camera pose match the fundamental physical truth. They proposed a framework for jointly training a single-view depth CNN and a camera pose estimation CNN from unlabeled video sequences, shown in Fig. \ref{fig:depth_and_pose_cnn_proposed_arch}. We adopted the network proposed by Bian \emph{et al.} \cite{bian2021unsupervised} to learn depth maps from monocular color images in our framework.

\section{METHODOLOGY}\label{methodology}

In this work, we propose a contrastive prioritized Deep-RL system that considers information from depth maps to perform the mapless navigation and obstacle avoidance of an UAV. 
% We train a network with a dataset collected on the environment to generate monocular depth image assumed as the input of the network CURL-based which can learn to control how to navigate these environments giving an adequate reward to this objective in a prioritized way.
We named our system of Depth-CUPRL, as this method is based on a CURL structure \cite{laskin2020curl} using depth maps and prioritized memories for navigation of an aerial vehicle. In the sequence, we explain our proposed approach and the other approaches used in this work for comparison. 

The Depth-CUPRL equation of motion is defined as:

\begin{equation}
v_t = f^{Depth-CUPRL}(I_t)
\label{eq:curldp}
\end{equation}
where $I_t$ is the raw pixel-colored monocular input image and $v_t$ is the velocity applied to the UAV.
For Depth-CUPRL, the Equation \ref{eq:curldp} passes through $Depth_{CNN}$ which generates a depth image of $I_t$, the CURL-based network extracts information from this depth map and then passes through a SAC-based network which gives the robot velocity. 
A schematic diagram of the Depth-CUPRL architecture is illustrated in Fig. \ref{fig:state_action} and pseudo-code can be
seen in Algorithm \ref{alg1}. It uses a \cite{bian2021unsupervised} dense block-based encoding network to generate depth maps. The KITTI dataset \cite{geiger2013vision} is used to train this network.

% %\subsection{Deep Reinforcement Learning}
% %
% %The Deep Reinforcement Learning problem could be summarized as controlling an agent in an environment trying to maximize its reward function. The deep Q-network (DQN) algorithm formulated in \cite{mnih2013playing}, \cite{mnih2015human} has been effective in many Atari video games. This algorithm was able to work at a human level using only a raw pixel image input to estimate agent action. Despite these achievements, DQN used only discrete action spaces to solve the complex observation spaces. However, when dealing with many tasks in robotics that involve control, it is necessary to make use of continuous action spaces. Therefore, applying another algorithm when leading continuous domain tasks is necessary.

% \subsection{Proposed Approach Structure of Depth-CUPRL}

% \cbox{ so aqui tu vai falar da tua metodologia, constroi a narrativa a partir da tua estrutura, e explica os detalhes da SAC, CURL e etc a partir disso, e nao o oposto!!! A essa altura o revisor cansou de ler teu paper e só quer rejeitar ele!}

%\begin{figure}[htbp!]
%\includegraphics[width=\linewidth]{images/fig4.pdf}
%\caption{Diagrama representando esquemática da Depth-CUPRL}
%\label{fig:curldp}
%\end{figure}

After training the Depth-CUPRL with the KITTI dataset, the network is \textit{finetuned} to improve depth estimation for the navigation environment. For this, images are extracted in sequences of random actions to improve the network's performance that generates the scene's depth. After that, the depth estimated is stacked in 3 consecutive temporal frames as the input from the encoder \textit{query} and \textit{key}, as shown in Figure \ref{fig:state_action}.
The images received in the encoders that extract information return the latent spaces $q$ and $k$, where $q$ is used in the Deep-RL network.
% The angular and linear velocity action applied to the robot is the network output.
The outputs of the Depth-CUPRL network are the linear and angular velocities are used to high-level control the UAV.

\begin{figure}[t]
\centering\includegraphics[width=\linewidth]{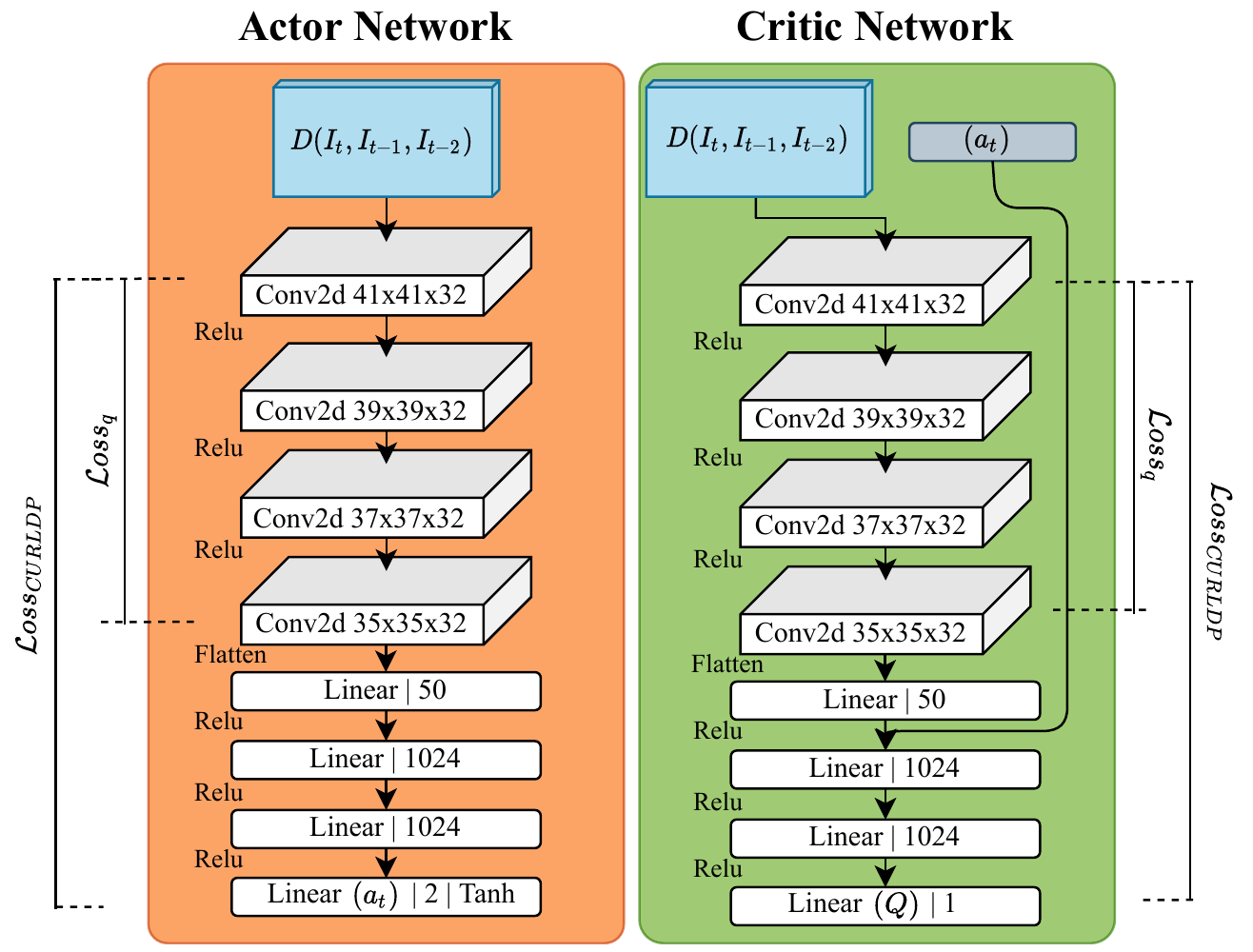}
\caption{Depth-CUPRL network architecture.}
\label{fig:arch}
\vspace{-5mm}
\end{figure}

The network architecture used by the Deep-RL resembles the SAC network presented in the works of de Jesus \emph{et al.} \cite{jesus2019deep}, and Grando \emph{et al.} \cite{grando2021deep}. Fig. \ref{fig:arch} shows the architecture including the \textit{query} encoders as convolutional layers. %The Depth-CUPRL network input has the current depth map and two previous agent estimates in the environment as the actor-network input.
The depth maps are processed by four convolutional layers and three fully connected layers. The number of layers and nodes are based on the proposal of Laskin \emph{et al.} \cite{laskin2020curl}. The output layer estimates angular and linear velocity to be used in the robot. However, the action value is normalized between $(-1;1)$ because of the hyperbolic tangent function $(tanh)$, which is the activation function. Therefore, the angular velocity and the linear velocity are also constrained between the ranges $ -0.25 $ to $0.25 $ $rad / s $ and $ 0 $ to $ 0.22 $ $ m / s $ , respectively.

\begin{algorithm}[hb]
    \algsetup{linenosize=\scriptsize}
    \scriptsize
    \caption{Depth-Imaged Contrastive Unsupervised Prioritized Representations in Reinforcement Learning}
    \label{alg1}
\begin{algorithmic}[1]
    \STATE Input: initial policy parameters $\phi$, Q-function parameters $\theta_1$, $\theta_2$, empty replay buffer $M$, maximum number of episodes $E$, maximum number of steps $T$
    \STATE Image network: depth estimation network trained $D$, contrastive encoder $\psi_q$ and $\psi_k$;\\
    \STATE Set target parameters equal to main parameters $\theta'_{1} \leftarrow \theta_1$, $\theta'_{2} \leftarrow \theta_2$
    \FOR{$episode=1$ until $E$}
        \STATE reset environment
        \FOR{$t=1$ until $T$}
            \STATE Observe image state $I_t$
            \STATE Image observation passed to depth map $d = D(I_t)$ 
            \STATE Depth map passed to $s = \psi_q(d)$
            \STATE Observe state $s$ and select action $a \sim \pi_{\phi}(\cdot|s)$
            \STATE Execute $a$ in the environment
            \STATE Observe next state $s'$, and reward $r$
            \STATE Store $(s,a,r,s',d)$ in replay buffer $M$
            \IF{update Reinforcement Learning}
                \STATE Randomly sample a batch of transitions, $B = \{(s,a,r,s')\}$ from $M$
                \STATE Compute targets for the Q functions: \\
                    \begin{equation*}
                        y (r,s') = r + \gamma\left(\min_{i=1,2} Q_{\theta'_{i}} (s', a') - \alpha \log \pi_{\phi}(a'|s')\right)
                    \end{equation*}
                \STATE Update the critic: \\
                \begin{equation*}
                    \nabla_{\theta_{i=1,2}} \frac{1}{|B|}\sum_{(s,a,r,s') \in B} \left( Q_{\theta_i}(s,a) - y(r,s') \right)^2
                \end{equation*}
                \STATE Update policy: \\
                \begin{equation*}
                    \nabla_{\phi} \frac{1}{|B|}\sum_{s \in B} \left(\min_{i=1,2} Q_{\theta_i}(s, a_{\phi}) - \alpha \log \pi_{\phi} \left(\left. a_{\phi} \right| s\right)\right)
                \end{equation*}
                \STATE Updates target:\\
                \begin{equation*}
                    \quad \theta'_{i=1,2} \leftarrow \tau \theta_{i} + (1-\tau) \theta_i
                \end{equation*}
            \ENDIF
            \IF{updade Constrative Learning}
                \STATE Randomly sample batch of depth map $d$, $B(d) = \{(s,a,r,s',d)\}$ from $M$\\
                \STATE Random augmentation $aug()$ sample for $query$ and $key$:\\
                        $\quad d_q = aug(d)$\\
                        $\quad d_k = aug(d)$
                \STATE Computes the latent space of $query$ and $key$:\\
                        $\quad q = \psi_q(d_q) $\\
                        $\quad k = \psi_k(d_k) $
                \STATE Updates $query$: \\
                    \begin{equation*}
                        \mathcal{L}oss_{\psi_q} = \log \frac{\exp(q^Tk_0)}{ \sum\nolimits_{i \in B(d)}^{}\exp(q^Tk_i) }
                    \end{equation*}
                \STATE Updates $key$:\\
                    \begin{equation*}
                        \quad {\psi_k} \leftarrow \tau {\psi_k} + (1-\tau){\psi_k}
                    \end{equation*}
            \ENDIF
        \ENDFOR
    \ENDFOR
\end{algorithmic}
\end{algorithm}
%\vspace{-7mm}

% \cbox{nao entendi nada desse paragrafo.} The critic network provides the Q-value for the current state and the agent's action, and the critic network has two copies that use the same amount of layers so that the actor network. The Q value is activated via a linear activation function.%:
%\begin{equation}
%y = kx + b
%\end{equation}
%where $ x $ is the input of the last layer, $ y $ is the Q value or the given predicted value of the current state, $ k $ is the trained weights $ b $ is the bias of the last layer.

\subsection{Reward Function}

%First, it is necessary to establish the reward and penalty system for the Deep-RL network.
The rewards and penalties system given to the agent is a simple one, based on empirical knowledge created during the problem-solving process.
Thus, the network does a feed-forward and back-propagation step to learn the hyperparameters. The reward function is formulated for the obstacle avoidance task, and it is defined as:
\begin{equation}
r (s_t, a_t) = 
\begin{cases}
r_{navigating} \ \textrm{if}\ min_x \geq c_o,
\\
r_{collision} \ \textrm{if}\ min_x < c_o,
\end{cases}
\end{equation}
where a negative reward $(r_{collision} = -1)$ is given if the robot collides with an obstacle. $ c_o $ is equivalent to the robot's distance of $62 cm $ from an obstacle or wall.
These collision conditions cause the agent to be reset to the initial training position.
Furthermore, $r_{navigating}$ is given when the robot is navigating the map, whose value was decided to be $0.01$.
The observation that the only thing we want the robot to avoid is colliding with objects or walls defined the rewards without creating a misleading reward for the agent to interpret the map \cite{sutton2018reinforcement}.

\subsection{Experimental Setup}
\label{chapter:vehicle_description}

\begin{figure}[tb]
    \centering
    \begin{subfigure}[b]{0.22\textwidth}
        \includegraphics[width=\textwidth]{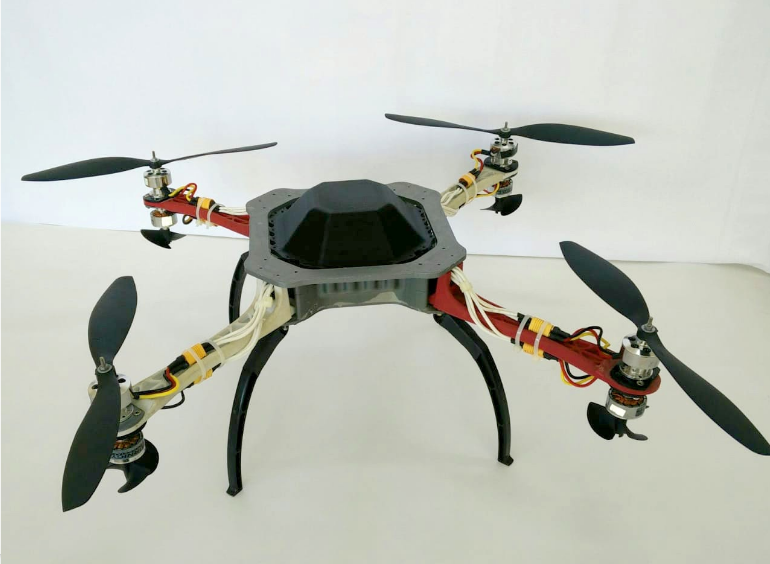}
        \caption{Real vehicle.}
        % \label{subfig:env1}
    \end{subfigure}
    \begin{subfigure}[b]{0.22\textwidth}
        \includegraphics[width=\textwidth]{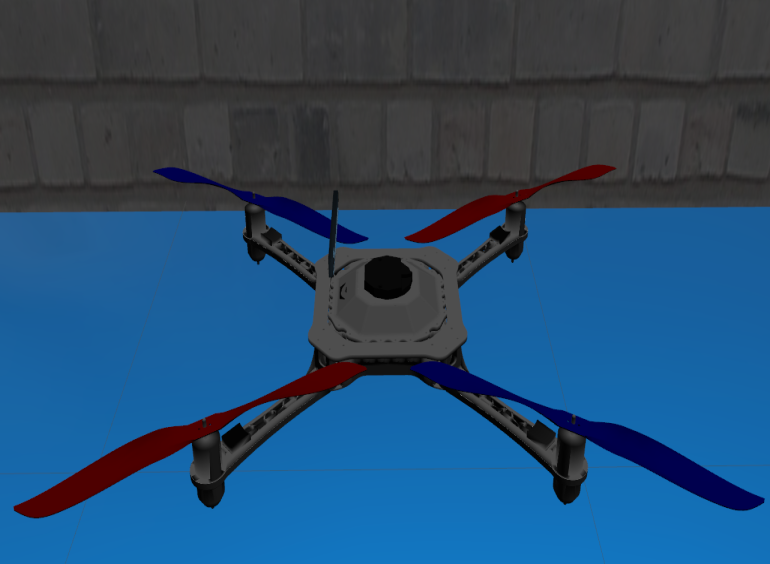}
        \caption{Simulated vehicle.}
        % \label{subfig:env2}
    \end{subfigure}
    \caption{Simulated vehicle based on the real Hydrone Vehicle.}
    \label{fig:vehicles}
    \vspace{-5mm}
\end{figure}

The simulations are run with the Robot Operating System (ROS) \cite{pyo2015ros} as backbone managing the connections between the proposed methods and the simulations. The chosen simulator is Gazebo for an easy connection with ROS and its realistic physics simulation. The algorithms are developed using the PyTorch library and OpenCV \cite{bradski2000opencv}. The robotic UAV platform used in the simulation is designed by Grando \textit{et al.} \cite{ricardo2021aprend}. The vehicle is based on a real vehicle called Hydrone \cite{drews2014hybrid,horn20}. The Hydrone vehicle is a quadrotor-like hybrid unmanned aerial underwater vehicle, but only the aerial capabilities are evaluated here. Fig. \ref{fig:vehicles} shows the real vehicle and the final vehicle described in simulation. The model in ROS can be controlled by linear and angular velocity in 6 degrees of freedom $(x, y, z, roll, pitch, yaw)$.

For the vehicle's base frame, the value of $1.93$ kg for mass is used. For the \textit{collision} of the frame, a box with dimensions of $0.47$ meters on a side and $0.09$ meters in height was estimated. The four rotors are placed symmetrically with an angle of $90^{\circ}$ between each and an exact distance of $0.27$ meters from the center of the vehicle. %In the Table \ref{table:hydrone} it is possible to see the parameters for the robot's performance.

%\begin{table}[H]
%\caption{Especificação feita no Hydrone}
%\begin{center}
%\begin{tabular}{l|l}
%\hline
%\textbf{Itens}                  & \textbf{Especificação} \\ \hline
%Velocidade translacional máxima & 0.22 m/s               \\ \hline
%Velocidade rotacional máxima    & 0.25 rad/s             \\ \hline
%Massa                           & 1.97                   \\ \hline
%\end{tabular}
%\label{table:hydrone}
%
%\end{center}
%\end{table}

% \begin{figure}[ht]
% \centering\includegraphics[width=13cm]{images/rexrov2.png}
% \caption{RexROV 2 Gazebo model.}
% \label{fig:rexrov2}
% \end{figure}

%\subsubsection{ROS and Gazebo}%NOT_DONE

%The robot operating system (ROS) is an open source robotics middleware, which acts as an intermediary, managing the connections between software and hardware in robots.
%ROS \cite{pyo2015ros} is not an operating system, but provides some standard operating system services such as inter-process messaging, low-level device control, hardware abstraction, and package management.

%It is really difficult to create an application for robotics without simulating it, so simulation is an essential tool in a roboticist's toolbox.
%Being able to design new robots and environments, test algorithms, collect data, apply artificial intelligence models easily and at the same time with realistic settings, all these characteristics are what makes a good simulator.
%On Gazebo, all the above features are possible \cite{fairchild2016ros}, and with the advantage of having an active community.
%Accompanies ROS in your installation with fully implemented bindings.

\subsection{Simulated Environments}

\begin{figure}[t]
    \centering
    \begin{subfigure}[b]{0.22\textwidth}
        \includegraphics[width=\textwidth]{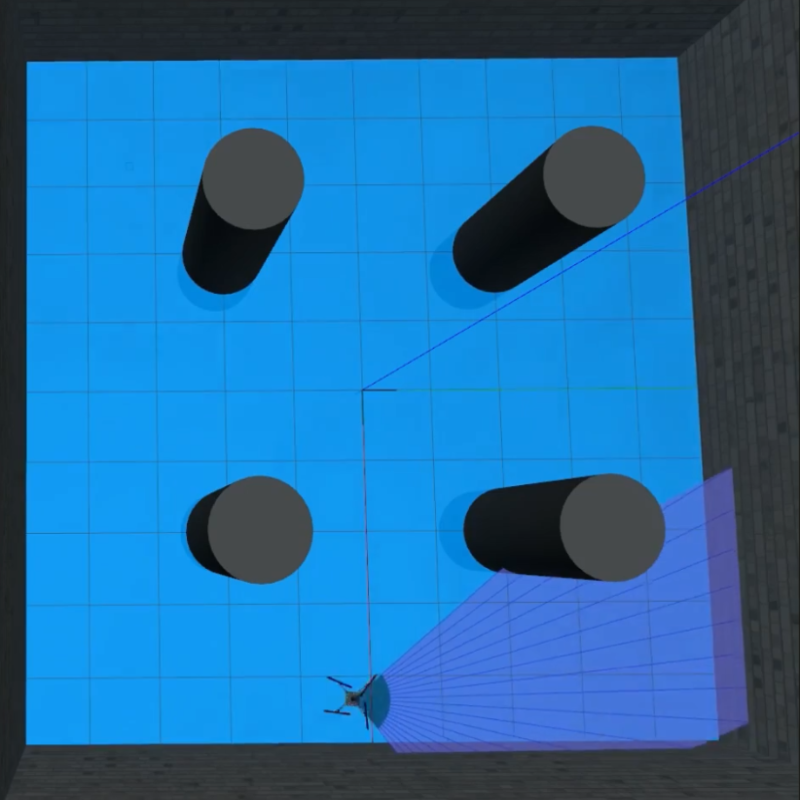}
        \caption{First environment.}
        \label{subfig:env1}
    \end{subfigure}
    \begin{subfigure}[b]{0.22\textwidth}
        \includegraphics[width=\textwidth]{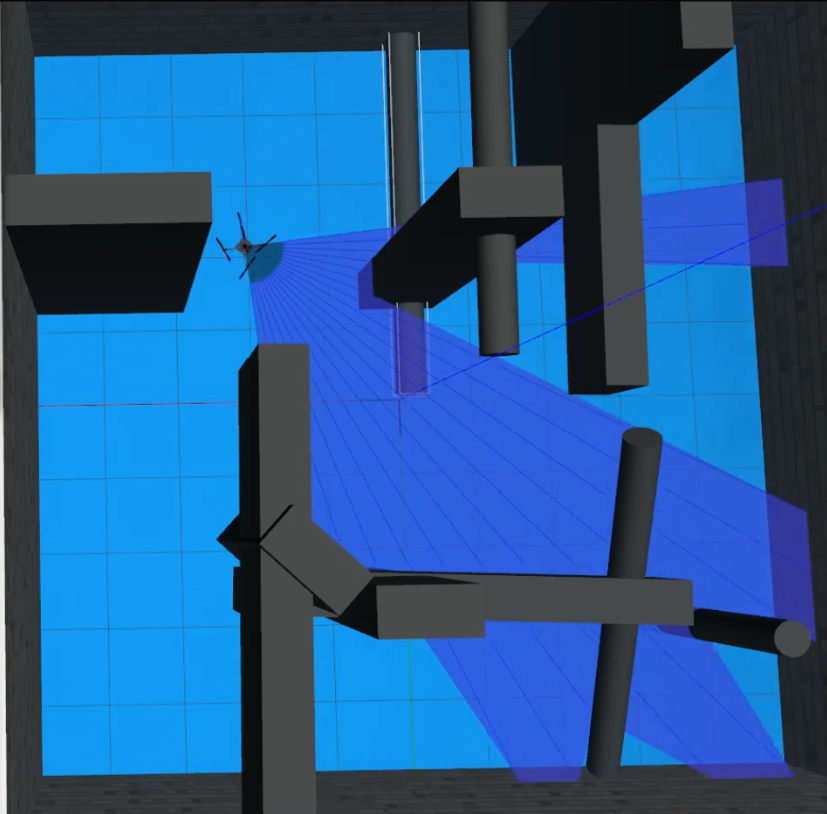}
        \caption{Second environment.}
        \label{subfig:env2}
    \end{subfigure}
    \caption{Simulated Environments used on Gazebo simulation.}
    \label{fig:env}
\vspace{-5mm}
\end{figure}

Both environments created in Gazebo can be seen in Figure \ref{fig:env}. The proposed environments are generated to show that the approach used in this work can navigate without colliding with walls and obstacles. In the first environment, four obstacles are allocated as illustrated in Figure \ref{subfig:env1}, the addition of these fixed obstacles produce a more challenging scenario for the robot. Therefore, if the vehicle collides with the wall or any obstacle, a negative reward is given for this action and the current episode stops. Fig. \ref{subfig:env2} show a more complex navigation scenario. This environment presents a higher degree of complexity than the first one since we add several obstacles.

\begin{figure*}[t]
  \subfloat[First environment.\label{fig:reward_1}]{
	\begin{minipage}[c][0.8\width]{0.5\linewidth}
	   \centering
	   \includegraphics[width=\linewidth]{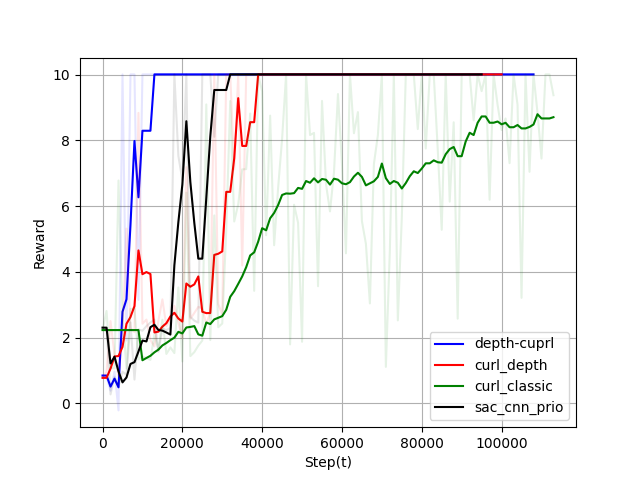}
	\end{minipage}}
 \hfill
  \subfloat[Second environment.\label{fig:reward_2}]{
	\begin{minipage}[c][0.8\width]{0.5\linewidth}
	   \centering
	   \includegraphics[width=\linewidth]{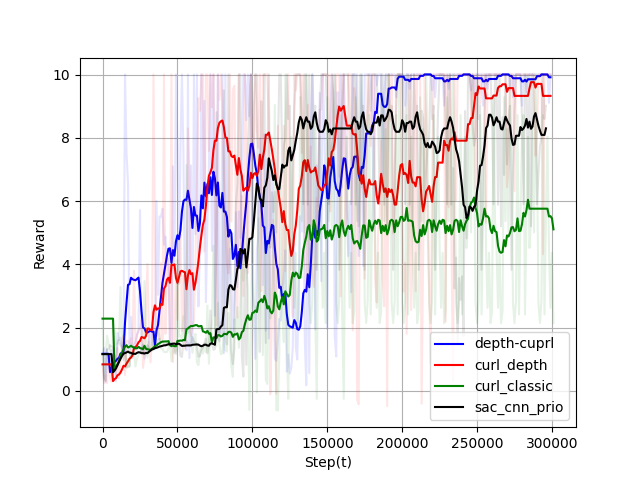}
	\end{minipage}}
 \hfill 
\label{fig:rewards}
\caption{Moving average of the reward over $100000$ and $300000$ training steps for the first and second environment, respectively.}
\vspace{-5mm}
\end{figure*}

\section{EXPERIMENTAL RESULTS}\label{results}

This section presents the discussion of the results.
%The data collected is related to the reward obtained from the project's artificial neural network.
Firstly, we present the reward gathered during the training. This information allows the estimation of the agent's degree of learning in the environment, as the improved in terms of reward is intrinsically linked to the agent's performance in the environment it must navigate.
All environments used for training the network are provided by the Grando \emph{et al.} \cite{ricardo2021aprend}. However, some changes are made in the source code of the Gazebo simulation to use the camera of the simulated UAV.

It is important to note that a Deep-RL agent is trained for each proposed environment. We compared our method Depth-CUPRL with the \emph{CURL} \cite{laskin2020curl} having as input the raw pixel image, and with an own version of the CURL called here \emph{CURL (Depth)} with depth maps as input but without prioritized memory. We also compared with a \emph{SAC (CNN prio.)} as adopted in Grando \textit{et al} \cite{grando2021deep} but with convolutional layers. The SAC (CNN prio.) uses depth maps as inputs of the network and has prioritized memory. All networks tested in the work followed the architecture proposed in Fig. \ref{fig:arch}. The initial position of the vehicle for both training and testing in each of the scenarios is defined as the Cartesian position (3.6, -2.6, 2.0) for the aerial environment. Only linear and angular velocities are applied to the $yaw$ of the vehicle.

Our method is configured with a prioritized replay memory of $35000$ samples for training in the first environment. Each episode has $1000$ steps $t$. The network is evaluated at each $1000$ step. The evaluation of the networks is carried out in such a way that it uses only the deterministic response of the network without any noise for $10$ episodes. With this, it is possible to evaluate the network's response by an average of tests and having as the maximum reward for evaluation of $10.00$.

The results of training the reward function of the first environment for the proposed and tested networks are shown in Figure \ref{fig:reward_1}. In the first episodes of training networks, negative rewards are noted. This condition is because the algorithm has started and is still learning. This episode reward means the robot is trying to maximize the task reward. All networks are trained for $100000$ steps. It is possible to conclude that the Depth-CUPRL, CURL (Depth), SAC (CNN prio.) networks successfully completed the proposed task in this environment. Depth-CUPRL presents higher efficiency to converge to the optimal reward. CURL based on RGB image \cite{laskin2020curl} remains unstable, failing to converge to an optimal result.

%\begin{figure}[t]
%\centering\includegraphics[width=\linewidth]{images/stage_1.png}
%\caption{Moving average of the reward over 100000 training episodes for the first environment}
%\label{fig:reward_1}
% \vspace{-5mm}
%\end{figure}

The results gathered in the second environment are shown in Fig. \ref{fig:reward_2}.
Given a higher complexity, a more significant number of training interactions is necessary for this environment. We adopted a replay memory of $140000$ and all methods are trained for almost $300000$ steps.
If compared with the previous reward functions, from Fig. \ref{fig:reward_1}, it is possible to notice a more unstable reward for the trained networks.
Despite this, the Depth-CUPRL obtains the highest average reward. Thus, we can see the advantage of using our technique in navigation and obstacle avoidance.

\section{DISCUSSION}\label{discussion}

\begin{table}[b]
\vspace{-5mm}
\caption{Comparison between our approach and other approach in the first environment. Bold letters highlights the best results.}
\begin{center}
\begin{tabular}{c c c c c}
\toprule
\textbf{}   \textbf{Algorithm}   & \textbf{Image}  & \textbf{Success} & \textbf{Crash} & \textbf{Success Rate (\%)} \\
\midrule
CURL (Depth)    &     Depth & \textbf{1000}             & \textbf{0 }             & \textbf{100\%}                  \\ 
CURL (Classic) &     Pixel        & 856              & 144            & 85.6\%                 \\ 
SAC (CNN prio.) &  Depth    & \textbf{1000 }            & \textbf{0  }            & \textbf{100\%} \\ \hline
Depth-CUPRL         &   Depth   & \textbf{1000 }            & \textbf{0 }             & \textbf{100\%  }                \\ 
\bottomrule
\end{tabular}
\label{table:env1}
\end{center}
%\vspace{-5mm}
\end{table}

After training in the first environment, a final model evaluation is performed after the 1000 episodes. Table \ref{table:env1} shows the results obtained for all methods. 

In this evaluation, it is possible to see that the algorithms that used information from depth images were able to complete the 1000 tests with 100\% success. Specifically, CURL (Classic), which uses pixel-based images, crashed many times in the evaluation. We can see in these results that the depth map images approach had a better efficiency in navigating through the first environment.

%\begin{figure}[t]
%\centering\includegraphics[width=\linewidth]{images/stage_2.png}
%\caption{Moving average of the reward over 300000 training steps for the %second environment}
%\label{fig:stage_2}
%\vspace{-5mm}
%\end{figure}

After training the approaches in the second environment, a final evaluation is performed at the end of the 1000 episodes. Table \ref{table:env2} shows the obtained results for all approaches.

\begin{table}[b]
\caption{Comparison between proposal and other networks in the second environment. Bold letters highlights the best results.}
\begin{center}
\begin{tabular}{c c c c c}
\toprule
\textbf{} \textbf{Algorithms}     & \textbf{Image}  & \textbf{Success} & \textbf{Crash} & \textbf{Success Rate (\%)} \\ 
\midrule

CURL (Depth)    &     Depth & 936             & 64              & 93.6\%                  \\ 
CURL (Classic)  &     Pixel        & 553            &   447          & 55.3\%                 \\ 
SAC (CNN prio.) &  Depth    & 822            &   178              & 82.2\%  \\\hline
Depth-CUPRL         &   Depth   & \textbf{992 }       & \textbf{8 }             & \textbf{99.2\% }                 \\ 
\bottomrule
\end{tabular}
\label{table:env2}
\end{center}
%\vspace{-5mm}
\end{table}

In this evaluation, the depth image approaches, collaborating with the results of Table \ref{table:env1}, had a better performance. Yet, none of the approaches were completed the scenario without a crash. Our approach Depth-CUPRL got the best performance crashing only 8 times of 1000, showing an overall better performance of 5.6\% in success rate. Only the approach that could not get a success rate above 80\% was the pixel-based method.

Taking into account information from depth images, we are able to increase the performance by 43.9\% in success rate in the best case (ours) and by 26.9\% in the worst depth-based method used by comparison (SAC (CNN prio)). 
\section{CONCLUSION AND FUTURE WORK}\label{conclusions}

In this work, we presented a novel Deep-RL based on the information of the estimated image depth of a simple monocular camera system to perform the mapless navigation and obstacle avoidance of UAVs. We evaluated our approach in two realistic scenario showing the Depth-CUPRL approach outperforms the state-of-the-art in terms of image-based RL approach. The use of depth maps is a commonplace in robotics and we show the advantage to use it in Deep-RL methods instead of raw pixel images.

With our approach, the generated depth maps are used as inputs in a network that learns these representations by image sequences. After that, this information is extracted and the Deep-RL agent trained to perform the vehicle's navigation and make it wise enough to avoid collisions with objects. The results show that the reward increases with the number of training episodes in the environment. 

The next steps will be focused in apply our approach also to the underwater navigation of the Hydrone \cite{drews2014hybrid, grando2021deep} using underwater depth maps \cite{drews2016underwater}. For these underwater scenarios, we intended to improve the depth estimation method to handle a high turbidity environment such as water and still be effective when in an aerial environment. Furthermore, real-world results are being conduced.
\section*{ACKNOWLEDGEMENTS}\label{acknowledgement}

The authors would like to thank the VersusAI team. This work was partly supported by the Technological University of Uruguay (UTEC), Federal University of Santa Maria (UFSM), Federal University of Rio Grande (FURG), and PRH-ANP.
\bibliographystyle{./bibliography/IEEEtran}
\bibliography{./bibliography/IEEEabrv,./bibliography/IEEEexample}

\end{document}